\documentclass[journal]{IEEEtran}

\usepackage{cite}
\usepackage{amsmath,amssymb,amsfonts}
\usepackage{graphicx}
\usepackage{textcomp}
\usepackage{booktabs}
\usepackage{siunitx}
\usepackage[section]{placeins}

\begin{document}

\title{Metallic Ultrasound Waveguides as a Distributed Tactile Sensing Platform}

\author{Alexandros~Rosakis, Alessio~Tamborini, Basile~Fakhoury, Cole~Bailey,
and Morteza~Gharib
\thanks{Alexandros Rosakis and Alessio Tamborini contributed equally to this work.
All authors were part of the Division of Engineering and Applied Science,
California Institute of Technology, Pasadena, CA 91125 USA
(Corresponding Author e-mail: ayrosaki@caltech.edu).}}

\maketitle

\begin{abstract}Tactile sensing is central to how robotic systems interact with the real
world, yet current solutions face a tradeoff between sensing area and system
complexity. This work investigates metallic ultrasound waveguides as
distributed tactile sensors fully interrogated from a single proximal
transducer. Using cylindrical indenters, we characterized the acoustic
response to single and multi-point contacts with varying forces and contact
materials. For single point indentation, the applied force was well captured
by a linear relationship with the ratio of the reflection to transmission
coefficients ($F = \alpha \cdot R/T$) across all nine tested materials
($R^{2} \geq 0.95$). The calibration slope, $\alpha$, correlated strongly with the
material's effective contact modulus (log--log Pearson $r = -0.98$). The
reflected energy partition was found to be a load-independent parameter
related to the contacting material's properties, enabling material
class differentiation independent of force. For the two-indenter experiment, both
contact forces were recovered from the waveguide signal and were in close
agreement with reference load cell measurements (contact~1, $R^{2} = 0.97$;
contact~2, $R^{2} = 0.95$). The approach was extended to two-dimensional
metallic sheets, confirming both contact localization and material-dependent
effects. Overall, these results validate metallic waveguides as a robust
platform for distributed tactile sensing, providing contact localization,
force estimation, and material-class discrimination for the contacting body.
\end{abstract}


\section{Introduction}
\label{sec:intro}

\IEEEPARstart{T}{he} sense of touch is an invaluable tool for how humans interact with the
world. Recently, there has been considerable research on how to grant this
ability to robotic systems, especially for dexterous manipulation of objects
and tasks such as robotic-assisted surgery~\cite{friebe_haptic_2026,
li_comprehensive_2024}. One major challenge is creating a sensor or sensor
array capable of measuring both force and its location on a robotic
manipulator's surface. Several different sensor types have been proposed.
Some common examples are capacitive sensing~\cite{ma_highly_2020,
li_research_2021}, piezoresistive sensing~\cite{phan_piezoresistive_2015},
piezoelectric sensing~\cite{shi_investigation_2014}, and magnetic
sensing~\cite{jamone_highly_2015}. Typically, these sensors are only capable
of point measurements and need to be arrayed to measure multiple separate
forces and their locations, which adds to overall system complexity. Optical
techniques have also been investigated. Some methods use cameras to image
contact gel layers that deform when touched~\cite{li_slip_2018,Li2026-sc}. This method
allows for spatial resolution force measurements, but suffers from a large form factor.
Optical Fiber Bragg Grating (FBG) techniques have also been
investigated~\cite{meribout_tactile_2024}. FBGs change their reflectance
spectra under mechanical stress, and several can be multiplexed onto a single
fiber. However, they require large, power-intensive FBG interrogators to
convert the light signals into usable data, which limits their use in robotic
systems.

More recently, acoustic and ultrasonic waveguides have received attention as
tactile sensors~\cite{li_disentangling_2025}. One implementation used a soft
polymer waveguide with embedded piezoelectric transducers, in which ultrasonic
wave packets are launched at one end and received at the
other~\cite{chossat_soft_2021}. Mechanical deformation alters the wave
propagation characteristics; for example, stretching the waveguide lengthens
the effective propagation length and increases the transit time between
emission and reception. The same principle has been applied for hand gesture
recognition using soft waveguides~\cite{alemu_echogest_2024}. In a pulse-echo
configuration, the contact location along the polymer waveguide can be measured
from the time-of-flight of the reflected wave packet. Liquid-filled acoustic
waveguides have also been proposed~\cite{lin_novel_2024}. Ye \textit{et al.}
\cite{ye_multipoint_2026} used a soft silicone waveguide with transducers at
both ends to simultaneously resolve two contact forces and locations. Beyond
soft media, acoustic waveguides have been integrated into structural elements
such as robot wheels for terrain and collision sensing~\cite{mason_acoustic_2024}.
While soft and liquid-filled waveguides are highly shape-compliant, the strong
attenuation of acoustic energy with propagation distance limits their achievable
sensing length. Acoustic methods have also been used to localize contact on
rigid waveguides~\cite{yang_adaptability_2021}. Ing \textit{et al.} used
acoustic time reversal to localize finger impact on a plate~\cite{ing_solid_2005}.
Bahrami \textit{et al.} used ultrasonic Lamb waves on touch glass to perform
contact localization via a machine learning approach~\cite{bahrami_machine_2022}.
Applications on stiff media exploit the low attenuation of acoustic waves to
enable larger-area sensing.

Metallic ultrasonic waveguides have a rich history of study. Cylindrical
metallic waveguides have been used for distributed temperature sensing in
extreme environments~\cite{periyannan_robust_2015}, measurements of viscous
liquid properties~\cite{vogt_measurement_2004}, soil loading and pressure
measurements~\cite{kurlenya_applicability_1997}, liquid level
sensing~\cite{melnikov_waveguide_1997, duan_research_2024}, and cure monitoring
of adhesives~\cite{vogt_cure_2003}. Compared with soft polymers and biological
materials, metals have higher wave speeds, lower acoustic attenuation, and 
higher acoustic impedance. This results in faster sampling, larger sensing
area, and minimal energy leakage to the surroundings. Furthermore, there are
several methods for exciting and measuring ultrasound waves in metallic
substrates, allowing for measurement flexibility. Despite these advantages, the
use of metallic waveguides for tactile sensing remains largely unexplored.

This study investigates the use of metallic ultrasound waveguides as
distributed tactile sensors. Three separate experiments were conducted. First,
single-point forces were applied to cylindrical metallic waveguides to explore
the interaction between force and wave propagation. Second, two simultaneous
and variable forces were applied to the waveguide to test distributed
force sensing. Finally, a planar metal waveguide was used to demonstrate the
parallel between the one-dimensional and the two-dimensional case.
\section{Methods}

\subsection{Transduction Mechanisms for Ultrasonic Waveguides}

Thin carbon steel wires were used as ultrasound waveguides in this experiment. These wires had a diameter of 0.64 mm, a length of 914 mm, and was composed of a ASTM A228 alloy steel. The material properties are described in Supplement Section ~\ref{sec:Supplement B}. Magnetostrictive excitation was performed at
one end of the wire using coils made from thin enamel-coated wire and a bias
magnet~\cite{kropf_ultrasonic_2007}. Separate actuation and sensing coils were
used in the transducer design. The wire waveguides were electroplated with
nickel to increase the magnetostrictive coefficient. A multifunction oscilloscope
(Analog Discovery Pro ADP2230) was used for both waveform generation to excite 
ultrasound waves and signal acquisition to measure returning echoes. 
The sensing coil was connected to an ultrasonic preamplifier (Model 5678, Olympus). 
Data sampling was performed at a \SI{25}{\mega\hertz} sampling rate for a 
signal length of \SI{400}{\micro\second}.

\subsection{Single Point Force Measurements}

A testing stand supported the transducer to suspend the waveguide. 
A portion of the waveguide was placed onto a PLA backing plate connected to a 
load cell, which was attached to a rigid frame. A rail system was designed around 
the load cell for repeatable perpendicular contact between the cylindrical 
indenter and the waveguide; the indenter holder further serves as a platform 
for placement of weights, it had an approximate weight 
of \SI{50}{\gram}. The entire system (backing plate and rails) ensured the cylindrical 
indenter contacted only the waveguide. Nine different stock materials were 
sourced as cylindrical indenters, all materials had a 1/8th inch diameter, these 
included Acrylic, Aluminium, Ceramic Alumina, Copper, Garolite G-10/FR4, 
Quartz Glass, Nylon, 1045 Steel, and Tungsten Carbide. The material properties of the indentors are summarized in Supplement Section ~\ref{sec:Supplement C}. 

The electronics included an Arduino Uno, a load cell with a signal amplifier module, 
and a multifunction oscilloscope. The Arduino Uno served as the synchronization 
board in a master-slave communication architecture, sending a trigger to the 
oscilloscope to initiate each capture. The Arduino also served as the data-capture 
system for the load cell. The load cell used a standard Wheatstone Bridge configuration 
to measure loads up to a maximum of \SI{5}{\kilo\gram} and was connected to a signal 
amplifier (HX711) operating at \SI{80}{\hertz}. The system was configured so that each capture 
cycle consisted of 50 samples acquired at \SI{80}{\hertz}. Load cell calibration was performed 
using standard procedures to determine the gain factor, using calibration weights of 
\SI{500}{\gram}, \SI{1000}{\gram}, and \SI{2000}{\gram}.

The experimental procedure for single-point force tests used calibration weights 
to generate repeatable contact forces between the waveguide and the indenter. First, 
the load cell was tared with the backing plate, but without the wire lying on it. 
Then the wire would be laid on the backing plate, and a capture cycle was performed 
to compute the baseline. The indenter is then placed on the rails and contacted to 
the waveguide; a second capture cycle is performed. The load was then sequentially 
increased by \SI{200}{\gram} using calibration weights on the indenter’s platform; at each step, 
a capture cycle was performed. This process was repeated 10 times until the total load 
on the waveguide reached \SI{2050}{\gram}. Lastly, all load was 
removed from the waveguide, and another capture cycle was performed. This process 
was repeated with every indenter material. 

\subsection{Contact Localization}
The transducer and waveguide were fixed to a stand mounted onto a manual linear translation stage. This allowed for the waveguide to be moved in relation to a cylindrical indentor. The translation stage was moved in 20, \SI{1}{\milli\meter}  increments, and 50 samples were recorded as each step. Next, the translation stage was used to move the waveguide with increasing step sizes. The smallest step size was \SI{50}{\micro\meter}. Subsequent, steps were increased by \SI{50}{\micro\meter} until a step size of \SI{600}{\micro\meter} was reached. 50 samples were collected for each step size increase.  

\subsection{Sub 0.05 N Force Measurements}
The transducer and waveguide were fixed to a rotation stage. A steel cylinder was fixed to a scale with a resolution of \SI{0.01}{\gram} and an accuracy of \SI{0.1}{\gram}. The scale was positioned such that, when the stage was rotated, the waveguide would press into the metal cylinder. Care was taken to ensure that the waveguide always contacted the cylinder parallel to the scale's measurement axis. Measurements were taken for every \SI{0.25}{\gram} weight increment from \SI{0}{\gram} to \SI{5}{\gram}. 50 samples were collected per increment.   

\subsection{Energy Framework for Wave Propagation}

An analytical framework was developed to model wave propagation along a
1D waveguide and to recover the energy partition of multiple reflection
events from the set of coupled returning echoes. Each reflection event is
registered at the co-located emission/sensing site as a returning wave
packet with a characteristic amplitude and arrival time. The waveguide was
assumed lossless (no measurable attenuation builds as the wave propagates
along the wire), so that all unrecovered energy is accounted for at the
reflection sites.

\begin{figure*}[tbp]
\centering
\includegraphics[width=\textwidth]{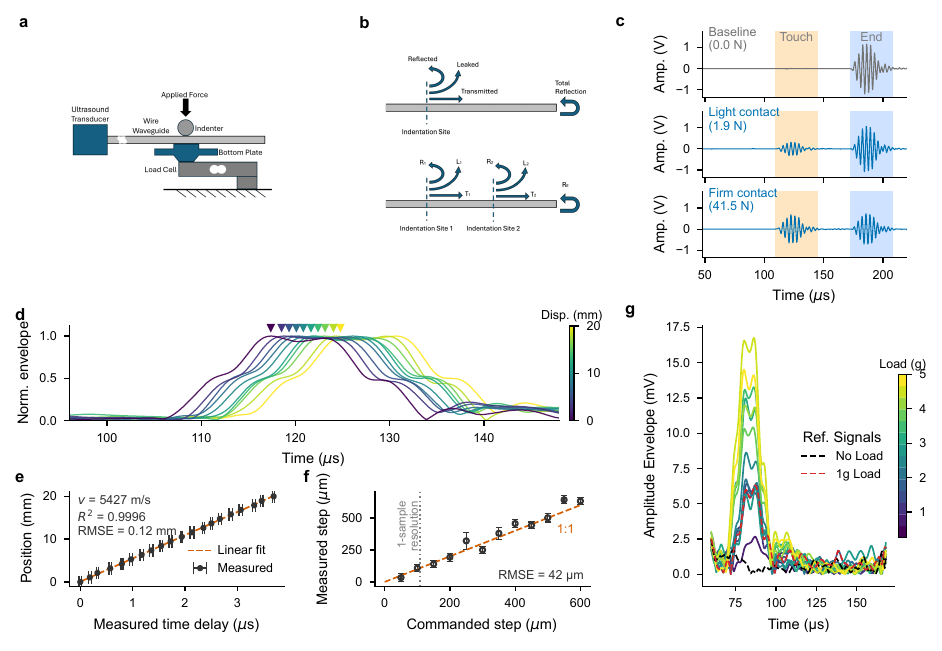}
\caption{Overview of ultrasound waveguides and experiment setup. (a) Shows the experimental setup comprising the ultrasound transducer, the wire waveguide, the loadcell and backing plate and the indenter. (b) shows a schematic of ultrasound wave propagation within a wire waveguide illustrating the energy partitions at each indentation site. (d) shows sample ultrasound signals at baseline, light contact, and firm contact. (e) shows the measured change in time of flight as an indentor was moved along the waveguide in set increments. (f) shows the measured change of the indentor as it was moved in increasing increments.(g) signal envelopes created by a steel contactor at \SI{0.25}{\gram} (\SI{0.0025}{\newton}) increments.}
\label{fig:framework}
\end{figure*}

Each contact along the waveguide produces a single partial reflection
(Fig.~\ref{fig:framework}b). At such event, the incident energy is
partitioned into reflected, transmitted, and lost fractions whose sum
equals unity,
\begin{equation}
R + T + L = 1
\label{eq:energy_conservation}
\end{equation}
where $R$, $T$, and $L \in [0, 1]$ are the reflection, transmission, and
loss coefficients, respectively, and each is expressed as a fraction of the
incident energy. The loss term $L$ groups all the energy that cannot be
recovered at the transducer level: energy that either leaks into the
indenter or undergoes mode conversion into waves that the transducer does
not measure. For incident energy $E_\text{in}$, a reflection event
therefore distributes as follows,
\begin{equation}
E_R = E_\text{in} R, \qquad
E_T = E_\text{in} T, \qquad
E_L = E_\text{in} L
\label{eq:energy_partition}
\end{equation}
in which $E_R$, $E_T$, and $E_L$ are the reflected, transmitted, and lost
energies, respectively. Considering the emission and sensing site are co-located at one
end of the waveguide, only returning waves are measured. On a bare wire
with no contacts, the incident waves propagate to the wire end and reflect
back. Reflections at the wire end are treated as perfect reflectors, that
is $R_\text{end} = 1$, $T_\text{end} = 0$, and $L_\text{end} = 0$. Hence,
we can define
\begin{equation}
\bar{E}_{\text{end},0} = E_\text{in} R_\text{end} = E_\text{in}
\label{eq:bare_wire}
\end{equation}
where $\bar{E}_{\text{end},0}$ denotes the measured energy at the sensor
location for the end reflection with no intermediate reflections. This
bare-wire echo serves as the incident energy reference. Introducing a
single contact produces at least two returning wave packets: one from the
contact and one from the end reflection. Tracking energy partition along
each path gives the following,
\begin{align}
\bar{E}_1 &= E_\text{in} R_1 \label{eq:single_contact} \\
\bar{E}_{\text{end},1} &= E_\text{in} T_1^2 R_\text{end}
\label{eq:single_end}
\end{align}
The end echo carries the squared transmission coefficient, $T_1^2$, because
it interacts twice with the contact point, once forward and once on return,
each time attenuating the signal by $T_1$. Generalizing, the energy
measured from the $n$-th reflection site along the waveguide is,
\begin{equation}
\bar{E}_n = E_\text{in} \left( \prod_{k=1}^{n-1} T_k^2 \right) R_n
\label{eq:general_energy}
\end{equation}
The above formulation states that the measured energy of reflection $n$
equals the incident energy reduced by the squared transmission coefficients
of every upstream contact, each traversed twice, and scaled by the
reflection coefficient of the $n$-th site, $R_n$. The index $n$ covers all
reflection sites encountered along the waveguide, ordered by wave arrival
time, and includes the wire end reflection site.

This mathematical framework assumes that all interactions generate distinct
peaks with respect to time of arrival and does not account for decoupling
superposed waves.

\subsection{Signal Analysis}
The signal measured using the transducer is a voltage--time trace of the
returning ultrasound wave packets. Voltage--time traces from the transducer
were filtered with a forward--backward (zero-phase) 4th-order Butterworth
bandpass filter (high-pass cutoff \SI{200}{\kilo\hertz}, low-pass cutoff
\SI{400}{\kilo\hertz}; sampling rate \SI{25}{\mega\hertz}). The signal
envelope was then obtained as the magnitude of the analytic signal computed
from the Hilbert transform of the real-valued trace. The peak envelope
amplitudes within a predefined region of interest were used as the
amplitude metrics. Regions of interest to isolate contact locations were
predefined using time of flight from sensor distance and longitudinal wave
speed.

Assuming the squared wave packet amplitude in the voltage domain is
proportional to the energy allows us to bridge the gap between the energy
framework and the measured wave signal,
\begin{equation}
A^2 \propto E
\label{eq:amp_energy}
\end{equation}
First, we quantify the incident wave energy from the amplitude of the end
reflection in the no-load instance,
\begin{equation}
\bar{E}_{\text{end},0} = E_\text{in} = A_{\text{end},0}^2
\label{eq:incident_ref}
\end{equation}
Using this as a baseline, we can solve for the single-point reflection and
transmission coefficients from Eqs.~\eqref{eq:single_contact} and
\eqref{eq:single_end} above, and $L$ can be subsequently quantified using
Eq.~\eqref{eq:energy_conservation},
\begin{equation}
\begin{aligned}
R_1 &= \frac{\bar{E}_1}{E_\text{in}} = \left( \frac{A_1}{A_{\text{end},0}} \right)^2, \\
T_1 &= \sqrt{\frac{\bar{E}_{\text{end},1}}{\bar{E}_{\text{end},0}}} = \frac{A_{\text{end},1}}{A_{\text{end},0}}, \\
L_1 &= 1 - R_1 - T_1
\end{aligned}
\label{eq:coefficients}
\end{equation}
The analysis then quantified the energy partition, defined as the ratio of
reflected energy to total extracted energy (the sum of reflected and lost
energy). A calibration model was subsequently established by linear
regression of the applied force against the ratio of reflected to
transmitted energy, constrained to pass through the origin, as follows,
\begin{equation}
F = \alpha \, \frac{R}{T}
\label{eq:calibration}
\end{equation}

\subsection{Distributed Force Measurements}

The experimental setup for the single-point force measurements was adapted
to add a second loading and measurement point to the rigid-body frame. This
involved adding an additional load cell (with a signal amplifier) and an
indenter rail system to the rigid-body frame, which was placed along the
path of the suspended waveguide. Both rail systems were preassembled with
steel indenters for this test. Both load cells were connected to the
Arduino for simultaneous logging of measurements at a trigger rate of
\SI{80}{\hertz}. The master--slave communication architecture between the
Arduino (master) and the oscilloscope (slave) was maintained. Capture
cycles were configured as previously described, in user-activated bursts of
50 samples.

The distributed point force measurement procedure involved varying the
loads applied to both indenters. The cumulative load applied on the waveguide 
was kept within \SI{20}{\newton}. The protocol first
applied loads to each indenter individually and then tested combinations of
loads across the two locations. Loads were applied as discrete steps using
calibration weights, with each step held while a burst of samples was
acquired. Combinatorial loading followed a mirrored pattern in which one
contact was held at a fixed level while the other was stepped through its
range, and vice versa, thereby isolating each contact's response and
enabling quantification of crosstalk.

The ultrasound waveguide signal was analyzed using the energy framework
presented above, extended to the case with two indenters. This analysis
decoupled the sequential contributions, extracting the energy partition
($R$, $T$, and $L$) at each reflection event, with the loss coefficient
$L$ providing the closure needed to resolve the two-contact system. The $R$
and $T$ coefficients were then mapped to the force at each indenter
location through the steel-indenter calibration model of
Eq.~\eqref{eq:calibration}, assuming the calibration applied uniformly
throughout the waveguide. Agreement between the estimated and measured
forces was quantified by the Pearson correlation coefficient ($r$) and the
root-mean-square error (RMSE), computed separately for each contact.
Estimation accuracy was further summarized by the coefficient of
determination ($R^2$) relative to the 1:1 identity line and by the
mean~$\pm$~standard deviation of the residual error.

\subsection{Two-Dimensional Waveguide Force Experiment}

Thin stainless-steel sheets (\SI{150}{\milli\meter} $\times$
\SI{150}{\milli\meter} $\times$ \SI{0.05}{\milli\meter}) were used as two-dimensional waveguides. Two
piezoelectric disks (SMD07T02R412WL, \SI{300}{\kilo\hertz} nominal resonant
frequency) served as combined transmitter--receiver elements and were
acoustically coupled to the sheet surface with a thin layer of
cyanoacrylate. A heat-set adhesive was applied over each disk to provide
mechanical damping. The two disks were placed at two adjacent edges of the 
sheet, with one disk taken as the coordinate origin (S2) and the other
at approximately $(145, 4)$~\si{\milli\meter} relative to it. 
A pulser--receiver (DPR300, BYK) operating in pulse-echo
mode excited each disk to generate a guided ultrasonic wave and then
captured and amplified the echoes received by the same disk. Signals were
digitized at a \SI{25}{\mega\hertz} sampling rate using the same
data-acquisition system described above. The two disks were excited and
recorded sequentially.

The first part of the experiment localized a point contact on the
waveguide. A cage structure was built to perform normal contact from an 
indenter in a rail system. The indenter was designed to hold a spherical 
ball of \SI{12.7}{\milli\meter} diameter at its tip.  The sheet rested on a flat PLA surface acting as a backing
plate. For each transducer, a reference signal was first acquired with no
applied load. A \SI{200}{\gram} calibration weight was placed on the indenters 
support structure sheet to generate a stable contact. Each disk was excited and recorded in turn.
The weight stack was applied at ten different locations across the plate,
and the corresponding contact coordinates were independently measured from
calibrated photographs for ground-truth reference.

For each transducer the loaded signal was subtracted from its no-load
reference to isolate the reflection produced by the contact. The arrival
time of the contact-induced wave was estimated from the difference signal
using an Akaike information criterion (AIC) onset picker applied within a
search window, with the window automatically expanded if the initial pick
fell too early. Arrival time was converted to distance using a wave
propagation speed of \SI{5450}{\meter\per\second} and accounting for
round-trip paths. The contact location was estimated by trilateration 
of the propagation lengths and the known transducer geometry. 
Localization accuracy was reported as the Euclidean distance between the 
estimated and ground-truth positions across the ten test points.

The second part of the experiment examined how the material in contact with
the waveguide influences the reflected signal. The indenter was placed at 
location four and same sized spheres of different materials were used in 
the indenter tip; the materials tested were stainless steel, aluminium, and PLA. For
each material, a no-load signal was acquired as a reference. The indenter was 
then placed on the sheet (~\SI{25}{\gram}) and another signal was captured.
Sequentially, the load was then increased in \SI{200}{\gram}
increments up to \SI{2000}{\gram}, with a signal recorded at each step. 
At each load level, the difference relative to
the no-load reference was computed. The amplitude of the reflected wave was
measured from the maximum of the difference signal within a fixed time
window and reported as a function of applied load for each material.
\section{Results}

\subsection{Single Point Force Measurements}
\begin{figure*}[tbp]
\centering
\includegraphics[width=\textwidth]{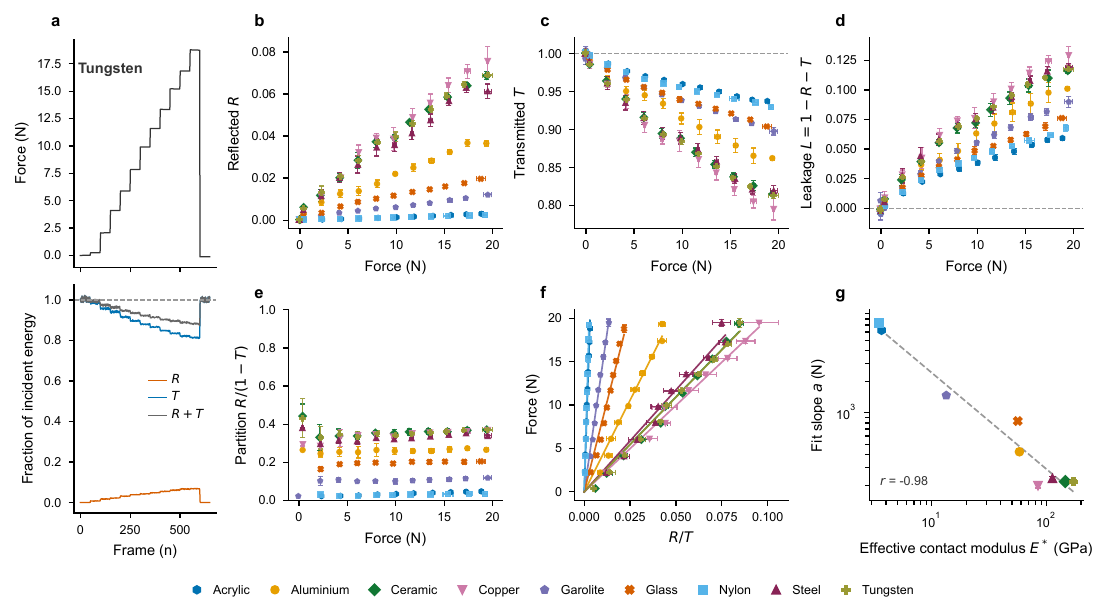}
\caption{Single indenter experiment. (a) shows a sample step wise loading curve and the reflection, R, and transmission, T,  analysis in the top and bottom plots, respectively. (b), (c), and (d) show the results of the energy framework analysis on the tested materials in the form of force versus reflection, transmission, and leakage, respectively. (e) plots the energy partition analysis with respect to indentation force. (f) plots the ratio of reflected to transmitted energy versus the applied force for all tested materials with a linear model fit. (g) log-log plot of the effective contact modulus between indenter and waveguide versus the slope of the F vs R/T fit. Error bars represent the standard deviation amongst experiment runs.}
\label{fig:single_point}
\end{figure*}
A single point of contact was shown to create a reflection event on the wire
(Figure  ~\ref{fig:framework}c), wherein a stronger contact generated a larger reflected
signal. Displacement of this indenter along the wire shifted the wave packet 
as show in Figure ~\ref{fig:framework}d ; a linearly proportional relationship was observed between
the measured time delay (single path time-of-flight) and the relative position
of the indenter. The proportionality constant was estimated using a linear regression
to \SI{5427}{\meter\per\second} (Figure~\ref{fig:framework}e), and corresponds to the wave propagation
speed in the waveguide. Figure~\ref{fig:framework}f shows that using algorithmic 
methods, subsample accuracy can be obtained on measured step size. Figure~\ref{fig:framework}g shows contact loads as small as \SI{0.005}{\newton} (\SI{0.5}{\gram}) create an observable signal peak.  

The single-point force experiment was performed on the carbon steel
waveguide using nine different indenter materials. 
Figure~\ref{fig:single_point}a shows example curves of the stepwise loading
for the single-point indentation (top) and the reflection/transmission
analysis results ($R$, $T$, and $R+T$) on the measured signal (bottom) for
tungsten, run~1. The $R+T$ sum falls below 1, indicating that energy is
lost into the indenter or converted to a different mode.
Figures~\ref{fig:single_point}b and~\ref{fig:single_point}c show the
computed $R$ and $T$ values, respectively, plotted against load force (in
newtons) for the nine different materials. As more force is applied, we
observe a monotonically increasing relationship for reflected energy $R$ and
a monotonically decreasing relationship for transmitted energy $T$.
Figure~\ref{fig:single_point}d plots the estimated lost energy $L$,
computed as $L = 1 - R - T$ from Eqn.~\eqref{eq:energy_conservation}, against the applied force, which also
increases monotonically with increasing load. Clear material effects are
observed in the $R$, $T$, and $L$ relationships with force.

\begin{table*}[!h]
\centering
\caption{Results of the $F$ versus $R/T$ fit for the single-indenter experiments.
$E^{*}$ is the effective contact modulus, $\alpha_{\mu}$ and $\alpha_{\text{SD}}$
are the mean and standard deviation of the fitted calibration slope,
$R^{2}$ is the coefficient of determination, RMSE is the root-mean-square
error, and \%FS is the percentage of full-scale span.}
\label{tab:calibration_fits}
\begin{tabular}{l S[table-format=3.1] S[table-format=4.0] S[table-format=3.0] S[table-format=1.2] S[table-format=1.1] S[table-format=1.1]}
\toprule
{Material} & {$E^{*}$ (\si{\giga\pascal})} & {$\alpha_{\mu}$ (\si{\newton})} & {$\alpha_{\text{SD}}$ (\si{\newton})} & {$R^{2}$} & {RMSE (\si{\newton})} & {RMSE (\%FS)} \\
\midrule
Nylon      & 3.5   & 7335 & 742 & 0.98 & 0.7 & 3.7 \\
Acrylic    & 3.6   & 6220 & 63  & 0.96 & 1.0 & 5.4 \\
Garolite   & 13.4  & 1469 & 51  & 0.98 & 0.8 & 4.3 \\
Glass      & 55.7  & 834  & 29  & 0.97 & 0.9 & 5.0 \\
Aluminium  & 57.5  & 422  & 10  & 0.96 & 1.0 & 5.2 \\
Copper     & 83.1  & 199  & 9   & 0.96 & 1.1 & 5.7 \\
Steel      & 110.5 & 235  & 8   & 0.95 & 1.2 & 6.0 \\
Ceramic    & 142.3 & 219  & 7   & 0.98 & 0.8 & 3.9 \\
Tungsten   & 167.1 & 218  & 9   & 0.99 & 0.6 & 3.2 \\
\bottomrule
\end{tabular}
\end{table*}
The extracted energy partition $R/(1-T)$, which represents the percentage
of reflected energy relative to extracted energy, appears to be a
force-independent material property, as indicated by the flat lines in
Figure~\ref{fig:single_point}e. Figure~\ref{fig:single_point}f plots the
ratio $R/T$ against the loading force and fits the calibration model
presented in Eq.~\eqref{eq:calibration} to each material's data. The fit
coefficient and results are reported in Table~\ref{tab:calibration_fits}.
As shown in Figure~\ref{fig:single_point}g, the fitted slope $\alpha$ of
the linearized force model (Eq.~\eqref{eq:calibration}) is inversely
correlated with the indenter's effective contact modulus, $E^{*}$, across
the nine materials, with a Pearson correlation $r = -0.98$ ($p < 0.05$) on
the log-transformed values.

\subsection{Distributed Force Measurements}
\begin{figure*}[!tbp]
\centering
\includegraphics[width=\textwidth]{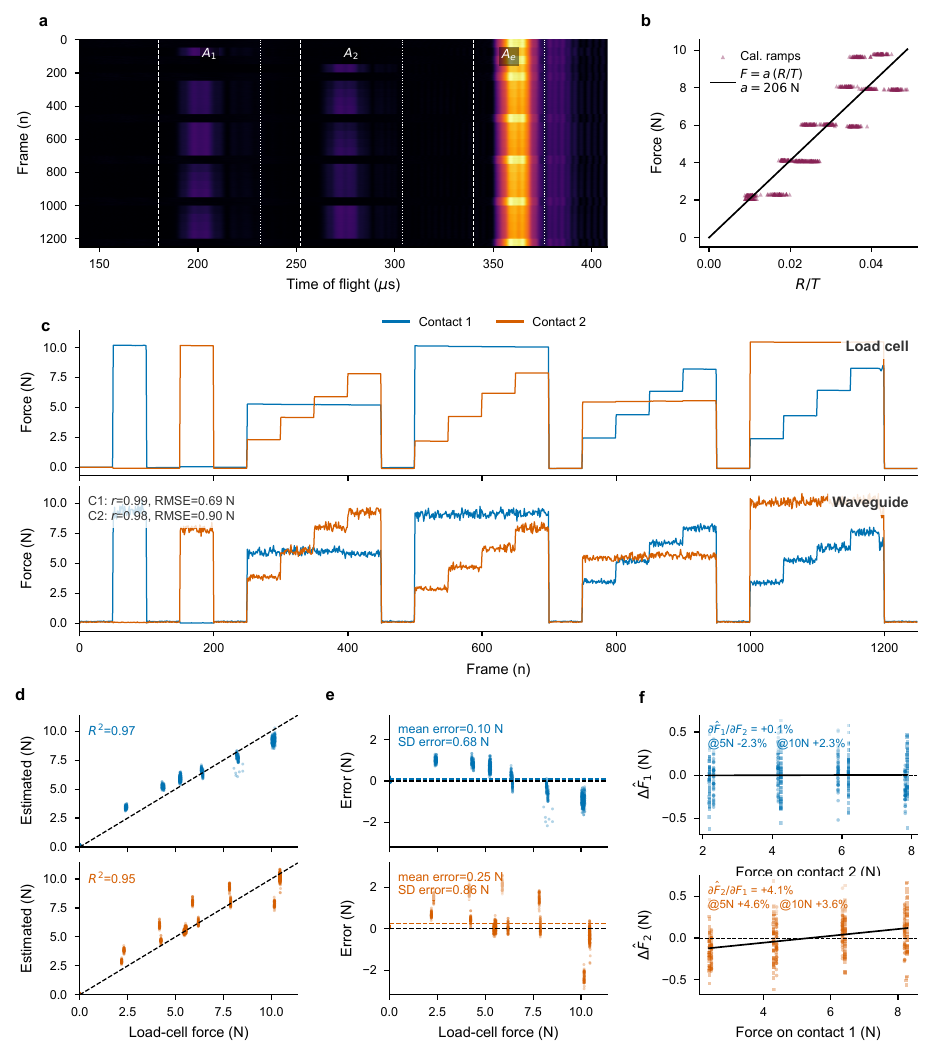}
\caption{Distributed force measurement on waveguide. (a) waterfall plot of the signal amplitude versus time of flight for all samples in the recording. Amplitude represented with magma color scale (b) shows the calibration curve of Force versus R/T for the steel indentor. (c) the dynamic force experiment traces showing the load cell measurements and the waveguide measurement in the top and bottom subplot, respectively. (d) shows the true versus estimated force separated for each contact point. (e) shows the residual (Estimated – True) analysis for the contact points. (f) shows the measurement coupling for the two contact point measurements on the waveguide.}
\label{fig:multipoint}
\end{figure*}

The metallic ultrasound waveguide was used for multipoint force
localization and measurement. The waveguide was loaded on a test rig with
two independent load cells and steel indenters; variable loads were applied
at the load cell locations, generating unique reflection patterns
(Figure~\ref{fig:multipoint}a). The steel-indenter force versus $R/T$ curve
in the \SIrange{0}{10}{\newton} range (Figure~\ref{fig:multipoint}b) is
used as the universal calibration for the waveguide on steel contact; the
curve yielded a calibration factor $\alpha = \SI{206}{\newton}$. The
coupled reflection and transmission analysis is performed for both contacts
to extract the $R/T$ coefficients and converted to force using
Eq.~\eqref{eq:calibration}. Figure~\ref{fig:multipoint}c shows the measured
force traces from the load cells (top) and the ultrasound waveguide
(bottom) for both contact points. Both contacts report an accurate force
estimate (contact~1, $r = 0.99$; contact~2, $r = 0.98$) and a low
associated root-mean-square error (contact~1, $\text{RMSE} = \SI{0.69}{\newton}$;
contact~2, $\text{RMSE} = \SI{0.90}{\newton}$). Ultrasound waveguide force
estimate accuracy is reported against the load cell (true) in
true-versus-predicted plots (Figure~\ref{fig:multipoint}d) and residual
plots (Figure~\ref{fig:multipoint}e) for both contacts; results show strong
linearity as measured with the coefficient of determination (contact~1,
$R^{2} = 0.97$; contact~2, $R^{2} = 0.95$) and minimal force
overestimation (contact~1, mean error $= \SI{0.10}{\newton}$; contact~2,
mean error $= \SI{0.25}{\newton}$). Figure~\ref{fig:multipoint}f depicts
measurement coupling between the two contact points on the waveguide,
showing no coupling for contact~1 ($+0.1\%$) and minimal coupling for
contact~2 ($+4.1\%$).

\subsection{Two-Dimensional Waveguide Force Experiment}
\begin{figure*}[!tbp]
\centering
\includegraphics[width=\textwidth]{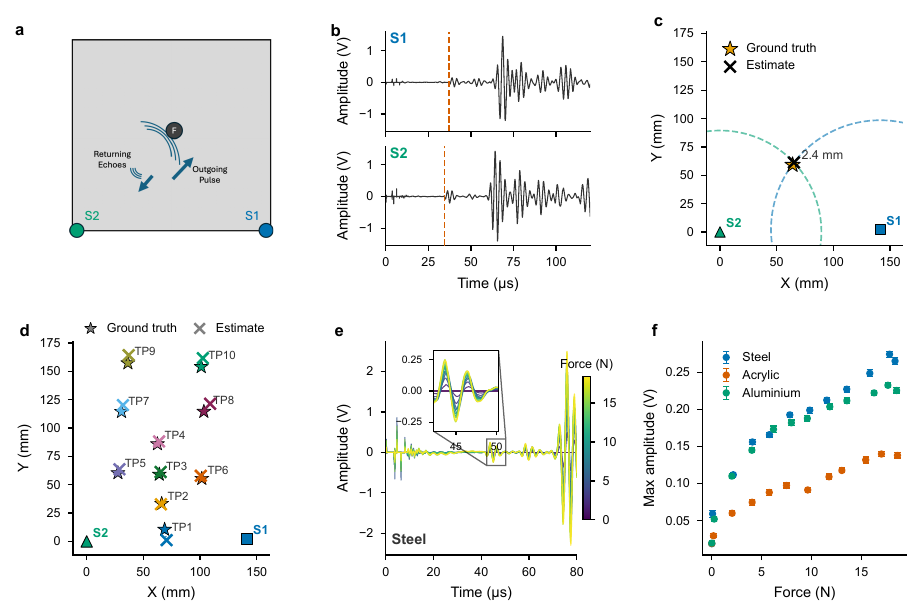}
\caption{Two-dimensional waveguide tactile sensing experiment. (a) shows the experimental setup comprising the sheet waveguide, the sensors and the weight. (b) shows sample traces of the two signals differences from reference with detected wave arrival time for S1 and S2. (c) Example contact localization on the two-dimensional sheet using trilateration on time of arrival from the two sensors. (d) Shows ten estimated versus ground truth localizations. (e) plots the signal from S1 with varying applied contact force. (f) shows the relationship between the maximum reflection amplitude and the applied force in newtons for different materials.}
\label{fig:twod}
\end{figure*}
The flat steel sheet was configured for tactile sensing using two
piezoelectric sensors as shown in Figure~\ref{fig:twod}a. Wave arrival time
was measured for each sensor as the signal difference with respect to the
baseline (Figure~\ref{fig:twod}b) and the longitudinal wave speed 
was then used to compute the intersection point via trilateration
(Figure~\ref{fig:twod}c). Figure~\ref{fig:twod}d shows the ground-truth
versus estimated positions for ten tested contact points. The method
reports an average localization error of 5.1 $\pm$ \SI{3.1 }{\milli\meter};
individual point results are summarized in Table~\ref{tab:localization}.
Qualitatively, the measured signal difference from contact shows a force
dependence, with the amplitude growing with increasing force
(Figure~\ref{fig:twod}e). In Figure~\ref{fig:twod}f, the relationship
between the maximum reflected amplitude and the applied force is examined
for different contact materials. As reported above the two-dimensional sheet
waveguide also displayed material-dependent effects for the force to amplitude relationship.

\begin{table}[tbp]
\centering
\caption{Two-dimensional sheet waveguide localization error ($n=10$).
RMSE is the root-mean-square error. SD is the standard deviation.}
\label{tab:localization}
\begin{tabular}{l c c S[table-format=4.1]}
\toprule
{Point} & {Ground truth} & {Estimated} & {Error} \\
      & {$(x, y)$ (\si{\milli\meter})} & {$(x, y)$ (\si{\milli\meter})} & {(\si{\milli\meter})} \\
\midrule
TP1  & $(68.7, 10.5)$   & $(70.5, 1.1 )$    & 9.6  \\
TP2  & $(66.1, 33.3)$   & $(65.5, 32.8  )$  & 0.8  \\
TP3  & $(64.0, 59.1)$   & $(65.2, 61.2 )$   & 2.4  \\
TP4  & $(62.4, 85.9)$   & $(63.8, 88.2 )$   & 2.7  \\
TP5  & $(27.5, 60.4)$   & $(28.9, 63.7 )$   & 3.5  \\
TP6  & $(101.5, 55.4)$  & $(100.6, 57.9 )$  & 2.6  \\
TP7  & $(30.7, 114.4)$  & $(32.0, 120.2 )$  & 5.9  \\
TP8  & $(103.3, 114.5)$ & $(108.9, 121.4 )$ & 8.9  \\
TP9  & $(36.3, 157.3)$  & $(37.0, 164.3 )$  & 7.0  \\
TP10 & $(101.0, 154.0)$ & $(102.4, 161.6 )$ & 7.8  \\
\midrule
\multicolumn{3}{l}{Mean $\pm$ SD (\si{\milli\meter})} & \multicolumn{1}{c}{$5.1 \pm 3.1$} \\
\multicolumn{3}{l}{RMSE (\si{\milli\meter})}          & 5.9  \\
\multicolumn{3}{l}{Max (\si{\milli\meter})}           & 9.6 \\
\bottomrule
\end{tabular}
\end{table}
\section{Discussion}
In this study, we investigated the use of metallic ultrasound waveguides as 
distributed tactile sensors, in which a proximal transducer interrogates contact 
along the waveguide's entire length via pulse-echo signal analysis. We 
demonstrated three main results: (1) analysis of the reflected and transmitted 
wave energy yields quantitative, distributed force measurements along the 
waveguide; (2) the material properties of the contacting body alter the 
acoustic signature upon contact in a load-invariant manner, enabling material
discrimination; and (3) the concept of waveguides as tactile sensors 
generalizes across dimensions, extending from wires to sheets. Together, 
these results demonstrate a versatile platform technology for contact sensing 
with a wide array of applications. 

The first part of this study focused on characterizing force-dependent
contact behaviors, as measured in the ultrasound signal, between our
cylindrical waveguide and a cylindrical indenter while undergoing crossed
contact. An ultrasound energy framework was developed to describe how
ultrasound waves interacted with different waveguide contact locations.
Increasing contact force consistently yielded a monotonically increasing
reflection coefficient, indicating that a larger proportion of the incoming
wave was being reflected. Interestingly, the material of the contacting
indenter strongly influenced the magnitudes of the reflection,
transmission, and loss coefficients at a given contact force, as clearly
shown in Figure~\ref{fig:single_point}b--d. The implications of this
finding are twofold. Firstly, it implies that for an accurate direct force
measurement from the waveguide, information about the contacting material
is necessary, as the slopes of the force versus $R/T$ curves for different
materials differ dramatically. This also allows different contact materials
to be used to control the waveguide's sensing range. Secondly, it suggests
an underlying material property modulates contact mechanics. Analyzing
the extracted energy partition, that is, the percentage of energy reflected
versus energy removed from forward propagation (i.e., reflected plus leaked
energy), revealed a clear material-dependent stratification. Most
importantly, the extracted energy-partition-versus-force curves stabilize at a constant value
at \SI{\sim2}{\newton} and remained at that value up to the maximum
measured force of \SI{\sim20}{\newton}. At lower force values
($F < \SI{2}{\newton}$), which correspond to the indenter's weight alone,
the observed noise in the data most likely originates from signal noise and
weak contact. This indicates that the energy partition parameter is a
load-independent metric that can discriminate between material classes
based on their properties. 

These findings are consistent with the broader principle established in 
guided-wave theory that mode energy is sensitive to the acoustic properties 
of the waveguide's boundary conditions. Prior work by Vogt et al.~\cite{vogt_measurement_2004, vogt_scattering_2003}
showed that for fully immersed waveguides, guided-wave modes leak energy into a surrounding viscous liquid and that the resulting attenuation depends on the acoustic properties of that liquid. Our results demonstrate that metallic waveguides subject to discrete, localized solid contacts produce reflection events whose characteristics depend on the material properties of the contacting body. This demonstrates that guided-wave interactions can be used 
to discriminate material properties from localized solid contact
extending the applicability of guided-wave material sensing to a fundamentally different contact regime.

The second part of this study built upon the single-point force analysis to
simultaneously measure two separate loads along the waveguide using a
single transducer at the proximal end. The primary challenge in measuring
distributed forces along the waveguide arises from the cascaded coupling of
sequential reflections; that is, how different contact locations interact
with the propagating ultrasound waves. For example, if two separate point
forces are applied to the waveguide, any force modulation to the contact
closer to the transducer will affect the amount of transmitted energy that
reaches the downstream locations. Therefore, the contributions of each
contact location must be calculated using the power framework presented
above before the force at each location can be estimated. While the wave
propagation analysis herein was performed for two contact points for this
experiment, it can be extended to an arbitrary number of points along the
wire.

In this experiment, to convert wave contributions to force, a single
calibration curve for a given contact material was applied to the entire
wire. This approach assumed a consistent calibration along the waveguide
and was foundationally grounded in the material homogeneity of the
waveguide along its length. The high correlations obtained in the results,
as presented in Figure~\ref{fig:multipoint}d, corroborate this assumption.
Yet, there are a couple of aspects of this approach worth discussing.
Firstly, we observe a non-linearity in the errors at the tested load
regimes (Figure~\ref{fig:multipoint}e) which are also observable in our
calibration model (Figure~\ref{fig:multipoint}b). This indicates that while
we simplified the calibration to a single linear parameter, a more complex
model is required to capture the full calibration curve. Secondly, we
observe greater variance in the error at the second contact point than at
the first, as illustrated in Figure~\ref{fig:multipoint}e. This observation
is in line with the expected coupling of wave reflections, which, as a
consequence, also introduces cumulative errors in downstream force
estimations. Nevertheless, upon applying the power framework to decouple
measured amplitudes to actual reflection and transmission coefficients, the
output at the second contact has only minimal ($\sim$4\%) coupling to the
first contact. Hence, the proposed analytical framework decouples the
relative contributions of each reflection for independent measurements.

The final part of this study extends the concept of a one-dimensional distributed-force
waveguide sensor to two-dimensional planar waveguides. We first explored contact-point
localization on the planar waveguide. As in the one-dimensional scenario,
time-of-flight can also be used to identify the contact location in two-dimensions.
Yet, given the additional dimension, multiple time-of-flight measurements
must be used to determine the position via trilateration. Our setup
consisted of the simultaneous analysis of pulse-echo signals from two
adhered piezoelectric transducers. In contrast to one-dimensional waveguides, where wave
propagation is confined to a single axis and yields a visually
interpretable signal, two-dimensional waveguides propagate waves radially, creating a
significantly more complex return pattern. Hence, to emphasize the arrival
of the reflected wave from contact, we analyzed the signal difference with
respect to the baseline (the signal with no contact). As shown in
Figure~\ref{fig:twod}b, this preprocessing method clearly demarcates the
wave arrival, allowing the computation of the contact location given the
wave speed (Figures~\ref{fig:twod}c and~\ref{fig:twod}d). The contact
location in this experiment was determined by placing an indentor with calibration weight
at each of the ten marked locations. The contact patch created by the spherical indentor created an ideal point contact that allowed for accurate location measurements. However, in practical use cases, not all contacts will be idealized point contacts. Large contact areas, such as from a large cylinder pressing evenly into the sheet, can introduce errors into the measurements. This limitation can be circumvented by designing sleeve or cover with strategically placed indentors. An example of this can be seen in Supplement Section ~\ref{sec:Supplement A} where a flexible sheet with incorporated spherical indentors was placed on a 2D waveguide. Beyond localization, we also explored
the force dependence of the signal reflection in sheet waveguides. This
confirmed the monotonically increasing relationship between applied force
and reflection amplitude, and testing different contact materials showed
similar signal characteristics as observed in one-dimensional waveguides. While these
results are more qualitative, they establish a parallel between one-dimensional and two-dimensional waveguides as tactile sensors.

The experimental results collected throughout this study lay the groundwork
for the design and implementation of metallic ultrasound waveguides as distributed
tactile sensors across a wide range of applications. Firstly, the
ultrasound waveguide platform technology uses a single sensor to interrogate
a large sensing area and can deconvolve multiple contact points for
independent readouts. Secondly, we observed that the contacting materials
on the bare waveguide significantly influence the coupling behavior.
Knowledge of this material effect can be used to design specific indentors, coatings, or sleeves, for both the 1D and 2D cases, to both modulate the sensor's operating range, which can vary along its length, and balance the trade-off between the number of
measurement points and the amount of reflected energy. Furthermore, the
material effect can also be exploited to gain information on the contacting material
in a load-independent manner. Given the robust yet flexible nature of
metallic waveguides, we see strong potential in integrating this technology
into robotic systems for enhanced tactile sensing that not only gives force
feedback but also has contacting material awareness. Future work will be
focused on the direct integration of this technology into robotic hands.

\section{Conclusion}
Several methods for distributed tactile sensing using metallic ultrasound waveguides have been presented. Firstly, this work presented and validated the use of ultrasound waveguides as distributed force sensors using a single proximal transducer. An energy-based framework was developed to model ultrasound wave propagation within the waveguide and decouple the effects of multiple sequential contact points. Application of this framework decouples contributions from individual measurement points and enables independent, accurate computation of force at each location. While two-point force measurements were shown, the method can be extended to an arbitrary number of points. Further, the findings provide evidence of load-independent material effects that can be used to measure the properties of the contacting materials. Finally, the method was extended to show feasibility in two-dimensional planar waveguides.

\section*{Acknowledgments}
During the preparation of this work the authors used Claude for proofreading, editing purposes and the generation of a 3D schematic concept image used in the supplemental documents. After using this tool/service, the authors reviews and edited the content as needed and take full responsibility for the scientific content and accuracy of the published article. 

\bibliographystyle{IEEEtran}
\bibliography{references}

@misc{mason_acoustic_2024,
	title = {Acoustic tactile sensing for mobile robot wheels},
	url = {http://arxiv.org/abs/2402.18682},
	doi = {10.48550/arXiv.2402.18682},
	publisher = {arXiv},
	author = {Mason, Wilfred and Brenken, David and Dai, Falcon Z. and Castillo, Ricardo Gonzalo Cruz and Cormier, Olivier St-Martin and Sedal, Audrey},
	month = feb,
	year = {2024},
}

@article{li_disentangling_2025,
	title = {Disentangling {Contact} {Location} for {Stretchable} {Tactile} {Sensors} from {Soft} {Waveguide} {Ultrasonic} {Scatter} {Signals}},
	volume = {7},
	issn = {2640-4567},
	url = {https://onlinelibrary.wiley.com/doi/abs/10.1002/aisy.202400561},
	doi = {10.1002/aisy.202400561},
	number = {5},
	journal = {Advanced Intelligent Systems},
	author = {Li, Zhiheng and Lin, Yuan and Shull, Peter B. and Ren, Hongliang},
	year = {2025},
	pages = {2400561},
}

@article{lin_novel_2024,
	title = {Novel, {Soft}, {Water}-{Filled} {Acoustic} {Waveguides} for {Simultaneous} {Tactile} {Force} and {Location} {Sensing}},
	volume = {71},
	issn = {1557-9948},
	url = {https://ieeexplore.ieee.org/document/10242239},
	doi = {10.1109/TIE.2023.3308140},
	number = {7},
	journal = {IEEE Transactions on Industrial Electronics},
	author = {Lin, Yuan and Shull, Peter B.},
	month = jul,
	year = {2024},
	pages = {8141--8155},
}

@article{alemu_echogest_2024,
	title = {{EchoGest}: {Soft} {Ultrasonic} {Waveguides} {Based} {Sensing} {Skin} for {Subject}-{Independent} {Hand} {Gesture} {Recognition}},
	volume = {32},
	issn = {1534-4320, 1558-0210},
	url = {https://ieeexplore.ieee.org/document/10556628/},
	doi = {10.1109/TNSRE.2024.3414136},
	journal = {IEEE Transactions on Neural Systems and Rehabilitation Engineering},
	author = {Alemu, Medhanit Y. and Lin, Yuan and Shull, Peter B.},
	year = {2024},
	pages = {2366--2375},
}

@article{ye_multipoint_2026,
	title = {Multipoint {Simultaneous} {Tactile} {Force} and {Location} {Sensing} {Through} {Soft} {Acoustic} {Waveguides}},
	volume = {75},
	issn = {0018-9456, 1557-9662},
	url = {https://ieeexplore.ieee.org/document/11522574/},
	doi = {10.1109/TIM.2026.3693449},
	journal = {IEEE Transactions on Instrumentation and Measurement},
	author = {Ye, Yufan and Lin, Yuan and Pan, Huiming and Sheng, Xinjun and Meng, Jianjun and Jiang, Shuo and Shull, Peter B.},
	year = {2026},
	pages = {9521712--9521712},
}

@article{periyannan_robust_2015,
	series = {Proceedings of the 2015 {ICU} {International} {Congress} on {Ultrasonics}, {Metz}, {France}},
	title = {Robust {Ultrasonic} {Waveguide} {Based} {Distributed} {Temperature} {Sensing}},
	volume = {70},
	issn = {1875-3892},
	url = {https://www.sciencedirect.com/science/article/pii/S1875389215007452},
	doi = {10.1016/j.phpro.2015.08.004},
	journal = {Physics Procedia},
	author = {Periyannan, S. and Rajagopal, P. and Balasubramaniam, K.},
	month = jan,
	year = {2015},
	pages = {514--518},
}

@inproceedings{kropf_ultrasonic_2007,
	title = {Ultrasonic magnetostrictive transducers for guided ultrasonic waves in thin wires},
	volume = {6532},
	url = {https://www.spiedigitallibrary.org/conference-proceedings-of-spie/6532/1/Ultrasonic-magnetostrictive-transducers-for-guided-ultrasonic-waves-in-thin-wires/10.1117/12.715815},
	doi = {10.1117/12.715815},
	booktitle = {Health {Monitoring} of {Structural} and {Biological} {Systems} 2007},
	publisher = {SPIE},
	author = {Kropf, Matthew M. and Tittmann, B. R.},
	month = apr,
	year = {2007},
	pages = {183},
}

@article{li_comprehensive_2024,
	title = {A comprehensive review of robot intelligent grasping based on tactile perception},
	volume = {90},
	issn = {0736-5845},
	url = {https://www.sciencedirect.com/science/article/pii/S0736584524000796},
	doi = {10.1016/j.rcim.2024.102792},
	journal = {Robotics and Computer-Integrated Manufacturing},
	author = {Li, Tong and Yan, Yuhang and Yu, Chengshun and An, Jing and Wang, Yifan and Chen, Gang},
	month = dec,
	year = {2024},
	pages = {102792},
}

@inproceedings{li_slip_2018,
	title = {Slip {Detection} with {Combined} {Tactile} and {Visual} {Information}},
	issn = {2577-087X},
	url = {https://ieeexplore.ieee.org/document/8460495/},
	doi = {10.1109/ICRA.2018.8460495},
	booktitle = {2018 {IEEE} {International} {Conference} on {Robotics} and {Automation} ({ICRA})},
	author = {Li, Jianhua and Dong, Siyuan and Adelson, Edward},
	month = may,
	year = {2018},
	pages = {7772--7777},
}

@article{bahrami_machine_2022,
	title = {Machine {Learning} for {Touch} {Localization} on an {Ultrasonic} {Lamb} {Wave} {Touchscreen}},
	volume = {22},
	issn = {1424-8220},
	url = {https://www.mdpi.com/1424-8220/22/9/3183},
	doi = {10.3390/s22093183},
	number = {9},
	journal = {Sensors},
	publisher = {Multidisciplinary Digital Publishing Institute},
	author = {Bahrami, Sahar and Moriot, J{\'e}r{\'e}my and Masson, Patrice and Grondin, Fran{\c c}ois},
	month = jan,
	year = {2022},
	pages = {3183},
}

@article{ing_solid_2005,
	title = {In solid localization of finger impacts using acoustic time-reversal process},
	volume = {87},
	issn = {0003-6951},
	url = {https://doi.org/10.1063/1.2130720},
	doi = {10.1063/1.2130720},
	number = {20},
	journal = {Applied Physics Letters},
	author = {Ing, Ros Kiri and Quieffin, Nicolas and Catheline, Stefan and Fink, Mathias},
	month = nov,
	year = {2005},
	pages = {204104},
}

@article{duan_research_2024,
	title = {Research on magnetostrictive liquid level gauge for water level measurement of steam generator},
	volume = {56},
	issn = {1738-5733},
	url = {https://www.sciencedirect.com/science/article/pii/S1738573324004285},
	doi = {10.1016/j.net.2024.08.049},
	number = {12},
	journal = {Nuclear Engineering and Technology},
	author = {Duan, Shuiqiang and Li, Minggang and Li, Jiaming and Li, Tongxi and Nie, Changhua and Yang, Zumao and Hu, Jun},
	month = dec,
	year = {2024},
	pages = {5422--5427},
}

@article{melnikov_waveguide_1997,
	title = {Waveguide ultrasonic liquid level transducer for nuclear power plant steam generator},
	volume = {176},
	issn = {00295493},
	url = {https://linkinghub.elsevier.com/retrieve/pii/S0029549397001556},
	doi = {10.1016/S0029-5493(97)00155-6},
	number = {3},
	journal = {Nuclear Engineering and Design},
	author = {Melnikov, V.I and Khokhlov, V.N},
	month = nov,
	year = {1997},
	pages = {225--232},
}

@article{yang_adaptability_2021,
	title = {Adaptability of {Ultrasonic} {Lamb} {Wave} {Touchscreen} to the {Variations} in {Touch} {Force} and {Touch} {Area}},
	volume = {21},
	issn = {1424-8220},
	url = {https://www.mdpi.com/1424-8220/21/5/1736},
	doi = {10.3390/s21051736},
	number = {5},
	journal = {Sensors},
	publisher = {Multidisciplinary Digital Publishing Institute},
	author = {Yang, Zengchong and Liu, Xiucheng and Wu, Bin and Liu, Ren},
	month = jan,
	year = {2021},
	pages = {1736},
}

@article{friebe_haptic_2026,
	title = {Haptic and {Palpation} {Sensing} for {Robotic} {Surgery}: {Engineering} {Perspectives} on {Design} and {Integration}},
	volume = {26},
	issn = {1424-8220},
	url = {https://www.mdpi.com/1424-8220/26/4/1126},
	doi = {10.3390/s26041126},
	number = {4},
	journal = {Sensors},
	publisher = {Multidisciplinary Digital Publishing Institute},
	author = {Friebe, Michael H.},
	month = jan,
	year = {2026},
	pages = {1126},
}

@article{ma_highly_2020,
	title = {Highly sensitive flexible capacitive pressure sensor with a broad linear response range and finite element analysis of micro-array electrode},
	volume = {6},
	issn = {2352-8478},
	url = {https://www.sciencedirect.com/science/article/pii/S2352847819301418},
	doi = {10.1016/j.jmat.2019.12.008},
	number = {2},
	journal = {Journal of Materiomics},
	author = {Ma, Longquan and Yu, Xuecheng and Yang, Yuanyuan and Hu, Yougen and Zhang, Xinyu and Li, Huayuan and Ouyang, Xing and Zhu, Pengli and Sun, Rong and Wong, Ching-ping},
	month = jun,
	year = {2020},
	pages = {321--329},
}

@article{li_research_2021,
	title = {Research progress of flexible capacitive pressure sensor for sensitivity enhancement approaches},
	volume = {321},
	issn = {0924-4247},
	url = {https://www.sciencedirect.com/science/article/pii/S0924424720317416},
	doi = {10.1016/j.sna.2020.112425},
	journal = {Sensors and Actuators A: Physical},
	author = {Li, Ruiqing and Zhou, Qun and Bi, Yin and Cao, Shaojie and Xia, Xue and Yang, Aolin and Li, Siming and Xiao, Xueliang},
	month = apr,
	year = {2021},
	pages = {112425},
}

@article{phan_piezoresistive_2015,
	title = {The {Piezoresistive} {Effect} of {SiC} for {MEMS} {Sensors} at {High} {Temperatures}: {A} {Review}},
	volume = {24},
	issn = {1941-0158},
	url = {https://ieeexplore.ieee.org/document/7243309},
	doi = {10.1109/JMEMS.2015.2470132},
	number = {6},
	journal = {Journal of Microelectromechanical Systems},
	author = {Phan, Hoang-Phuong and Dao, Dzung Viet and Nakamura, Koichi and Dimitrijev, Sima and Nguyen, Nam-Trung},
	month = dec,
	year = {2015},
	pages = {1663--1677},
}

@article{shi_investigation_2014,
	title = {Investigation on {Multi}-{Piezoelectric} {Effects} from the {First} {Positive} {Piezoelectric} {Effect}},
	volume = {609-610},
	issn = {1662-9795},
	url = {https://www.scientific.net/KEM.609-610.1398},
	doi = {10.4028/www.scientific.net/KEM.609-610.1398},
	journal = {Key Engineering Materials},
	publisher = {Trans Tech Publications Ltd},
	author = {Shi, Li Ping and Huang, Jie and Wei, Xi Wen and Wei, Yan Bo},
	year = {2014},
	pages = {1398--1403},
}

@article{jamone_highly_2015,
	title = {Highly {Sensitive} {Soft} {Tactile} {Sensors} for an {Anthropomorphic} {Robotic} {Hand}},
	volume = {15},
	issn = {1558-1748},
	url = {https://ieeexplore.ieee.org/document/7070742},
	doi = {10.1109/JSEN.2015.2417759},
	number = {8},
	journal = {IEEE Sensors Journal},
	author = {Jamone, Lorenzo and Natale, Lorenzo and Metta, Giorgio and Sandini, Giulio},
	month = aug,
	year = {2015},
	pages = {4226--4233},
}

@article{meribout_tactile_2024,
	title = {Tactile sensors: {A} review},
	volume = {238},
	issn = {0263-2241},
	url = {https://www.sciencedirect.com/science/article/pii/S026322412401217X},
	doi = {10.1016/j.measurement.2024.115332},
	journal = {Measurement},
	author = {Meribout, Mahmoud and Abule Takele, Natnael and Derege, Olyad and Rifiki, Nidal and El Khalil, Mohamed and Tiwari, Varun and Zhong, Jing},
	month = oct,
	year = {2024},
	pages = {115332},
}

@article{chossat_soft_2021,
	title = {Soft {Acoustic} {Waveguides} for {Strain}, {Deformation}, {Localization}, and {Twist} {Measurements}},
	volume = {21},
	issn = {1558-1748},
	url = {https://ieeexplore.ieee.org/document/9152997},
	doi = {10.1109/JSEN.2020.3013067},
	number = {1},
	journal = {IEEE Sensors Journal},
	author = {Chossat, Jean-Baptiste and Shull, Peter B.},
	month = jan,
	year = {2021},
	pages = {222--230},
}

@article{vogt_measurement_2004,
	title = {Measurement of the material properties of viscous liquids using ultrasonic guided waves},
	volume = {51},
	issn = {1525-8955},
	url = {https://ieeexplore.ieee.org/document/1320854},
	doi = {10.1109/TUFFC.2004.1304272},
	number = {6},
	journal = {IEEE Transactions on Ultrasonics, Ferroelectrics, and Frequency Control},
	author = {Vogt, T.K. and Lowe, J.S. and Cawley, P.},
	month = jun,
	year = {2004},
	pages = {737--747},
}

@article{kurlenya_applicability_1997,
	title = {Applicability of acoustic waveguides for stress measurement in soils},
	volume = {33},
	issn = {1573-8736},
	url = {https://doi.org/10.1007/BF02765435},
	doi = {10.1007/BF02765435},
	number = {1},
	journal = {Journal of Mining Science},
	author = {Kurlenya, M. V. and Petrov, V. E. and Popov, S. N. and Tkach, Kh. B.},
	month = jan,
	year = {1997},
	pages = {88--93},
}

@article{vogt_cure_2003,
	title = {Cure monitoring using ultrasonic guided waves in wires},
	volume = {114},
	issn = {0001-4966},
	url = {https://doi.org/10.1121/1.1589751},
	doi = {10.1121/1.1589751},
	number = {3},
	journal = {The Journal of the Acoustical Society of America},
	author = {Vogt, T. and Lowe, M. and Cawley, P.},
	month = aug,
	year = {2003},
	pages = {1303--1313},
}

@article{vogt_scattering_2003,
	title = {The scattering of guided waves in partly embedded cylindrical structures},
	volume = {113},
	issn = {0001-4966},
	url = {https://doi.org/10.1121/1.1553463},
	doi = {10.1121/1.1553463},
	number = {3},
	journal = {The Journal of the Acoustical Society of America},
	author = {Vogt, T. and Lowe, M. and Cawley, P.},
	month = feb,
	year = {2003},
	pages = {1258--1272},
}

@ARTICLE{Li2026-sc,
  title     = "Biomimetic multimodal tactile sensing enables human-like robotic
               perception",
  author    = "Li, Shoujie and Wu, Tong and Xu, Jianle and Huang, Yan and
               Zhang, Zongwen and Zhao, Hongfa and Xu, Qinghao and Wang, Zihan
               and Ye, Linqi and Yang, Yang and Lyu, Chuqiao and Mu, Shilong
               and Wang, Xueqian and Xie, Zhaoqian and Wu, Changsheng and Yu,
               Xinge and Ding, Wenbo",
  journal   = "Nat. Sens.",
  publisher = "Springer Science and Business Media LLC",
  volume    =  1,
  number    =  1,
  pages     = "52--62",
  month     =  jan,
  year      =  2026,
  copyright = "https://www.springernature.com/gp/researchers/text-and-data-mining",
  language  = "en"
}

\clearpage

\section{Supplemental Information}
\subsection{2 Dimensional Indentor Sleeve}
\label{sec:Supplement A}
\begin{figure*}[tbp]
\setcounter{figure}{0} 
\renewcommand{\thefigure}{S\arabic{figure}} 
\centering
\includegraphics[width=\textwidth]{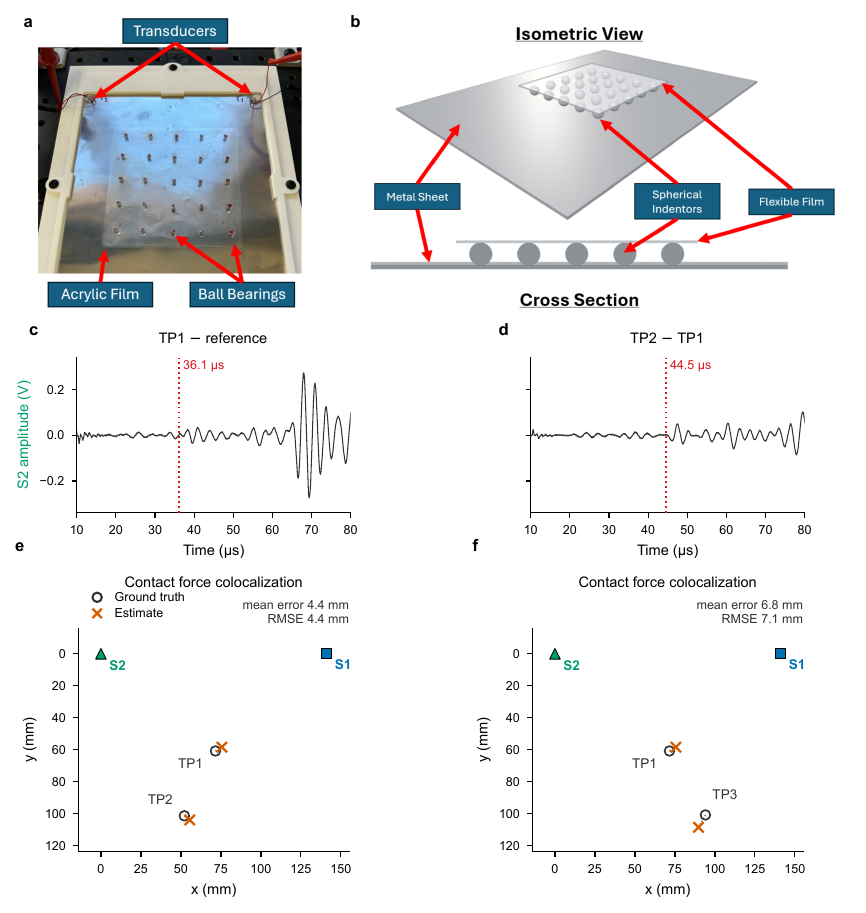}
\caption{Indentor Sleeve on 2 Dimensional Waveguide a) Image showing an indentor sleeve made of an acrylic sheet and steel ball bearings on the 2 dimensional waveguide. b) 3D rendering showing the indentor sleeve concept. c) Example signal trace from when a 200 gram weight was placed at TP1 showing echo arrival time. Note that the signal was referenced by subtracting the signal when no force was applied to the indentor sheet. d) Example signal trace from when a 200 gram weight was placed at TP2 showing echo arrival time. Note that a weight was still present at TP1 and therefore, the signal was referenced by subtracting the raw signal from TP1 in order to remove the baseline echoes as well as the contribution from the force at TP1. e) Estimated versus ground truth for TP1 and TP2 when they were both pressing into the waveguide showing a mean accuracy of 4.4 mm. f) Estimated versus ground truth for TP1 and TP3 when they were both pressing into the waveguide showing a mean accuracy of 6.8mm. }
\label{fig:Appendix}
\end{figure*}

We propose a method to remove contact material and contact patch ambiguity when using 2 dimensional metal sheets as distributed force sensors. Our method involves creating a flexible sheet or sleeve adhering several large number of spherical indentors to its surface in such a way that, when the flexible sheet or sleeve is placed on to the metal sheet, only the indentors come into contact with the metal surface. This way, when force is applied to the flexible sheet, the spherical indentors closest to the contact force will be pressed into the metal sheet and create reflections. The spherical indentors can either be uniformly or randomly distributed onto the flexible sheet. Figure ~\ref{fig:Appendix} b shows an example 3-D rendering of one such sheet.

A demonstration of such a sheet or sleeve was created using a thin acrylic sheet as the flexible substrate. \SI{2}{\milli\meter} steel ball bearing were adhered to the acrylic sheet using cyanoacrylate adhesive. They were placed in a 5 by 5 grid on the acrylic sheet with a spacing of \SI{2}{\centi\meter} between each column and row. The same 2 dimensional metal waveguide was used as in the Two-Dimensional Waveguide Force Experiment. The flexible indentor sheet was placed onto the metal waveguide surface. Figure ~\ref{fig:Appendix} a shows an image of the indentor sheet on the metal waveguide. Reference signals were taken for each piezoelectric transducer while no force was applied to the indentor sheet. Then a \SI{200}{\gram} calibration weigh was placed at the central indentor position, TP1 (row 3, column 3 where column number increases from left to right and row number increases from bottom to top). Signals from each transducer were then captured. A second 200 gram weight was added to position TP2 (row 2, column 2) and more signals were recorded from each transducer. Finally, the weight was removed from TP2 and placed at TP3 (row 2, column 4) and signals from the transducers were recorded. In the case of the single weight at TP1, the echo arrival times were isolated by subtracting the reference signals from the TP1 signals. Figure ~\ref{fig:Appendix} c shows an example waveform for point TP1 after referencing. In the case of the TP2 and TP3 signals, the TP1 signal was used as a reference to isolate the echoes from these further points. Figure ~\ref{fig:Appendix} d shows an example waveform after the TP1 signal has been subtracted for TP2. Wave arrival times for each point were then measured. Trilateration was used to estimate their position using the longitudinal wave speed in steel. The estimated positions relative to their ground truth positions can be seen in figure~\ref{fig:Appendix} e and f. Note that the position of TP2 and TP3 were measured when a force was applied at TP1 showing that separate force locations can be measured simultaneously.  

\clearpage
\subsection{Waveguide Properties} 
\label{sec:Supplement B}
\begin{table}[!h]
\setcounter{table}{0} 
\renewcommand{\thetable}{S\arabic{table}} 

\centering
\caption{Structural properties of the waveguide.}
\label{tab:waveguide_props}
\begin{tabular}{ll}
\toprule
Property & Value \\
\midrule
Material                        & Carbon steel \\
Alloy                           & ASTM A228 \\
Shape                           & Wire \\
Diameter                        & \SI{0.64}{\milli\meter} \\
Length                          & \SI{914}{\milli\meter} \\
Density, $\rho$                 & \SI{7850}{\kilogram\per\cubic\meter} \\
Longitudinal wave speed, $c_L$  & \SI{5900}{\meter\per\second} \\
Young's modulus, $E$            & \SI{210}{\giga\pascal} \\
Poisson's ratio, $\nu$          & 0.313 \\
\bottomrule
\end{tabular}
\end{table}
\subsection{Indentor Properties}
\label{sec:Supplement C}
\begin{table}[!h]
\setcounter{table}{1} 
\renewcommand{\thetable}{S\arabic{table}} 
\centering
\caption{Material properties of the cylindrical indenters used for force testing.}
\label{tab:indenter_props}
\sisetup{table-number-alignment = center}
\begin{tabular}{l S[table-format=3.1] S[table-format=1.2] S[table-format=5.0] S[table-format=5.0]}
\toprule
{Material} & {$E$ (\si{\giga\pascal})} & {$\nu$} & {$\rho$ (\si{\kilogram\per\cubic\meter})} & {$c_L$ (\si{\meter\per\second})} \\
\midrule
Acrylic             & 3.2 & 0.37 & 1190  & 2730  \\
Aluminium           & 69  & 0.33 & 2700  & 6350  \\
Ceramic Alumina     & 372 & 0.22 & 3875  & 10520 \\
Copper              & 117 & 0.34 & 8912  & 5010  \\
Garolite G-10/FR4   & 14  & 0.12 & 2050  & 3000  \\
Glass Quartz        & 72  & 0.17 & 2214  & 5740  \\
Nylon               & 3   & 0.39 & 1140  & 2600  \\
1045 Steel          & 200 & 0.29 & 7850  & 5900  \\
Tungsten Carbide    & 630 & 0.21 & 15000 & 7000  \\
\bottomrule
\end{tabular}
\end{table}

\end{document}